%% file: main.tex
\documentclass[10pt,twocolumn,letterpaper]{article}

\usepackage{amsmath}
\usepackage{amssymb}
\usepackage{mathtools}
\usepackage{amsthm}

\newtheorem{theorem}{Theorem}
\usepackage[pagenumbers]{iccv} 

\definecolor{iccvblue}{rgb}{0.21,0.49,0.74}
\usepackage[pagebackref,breaklinks,colorlinks,allcolors=iccvblue]{hyperref}

\def\paperID{4444} 
\def\confName{ICCV}
\def\confYear{2025}

\title{\texttt{Dig2DIG}: Dig into Diffusion Information Gains for Image Fusion}

\author{
\textbf{Bing Cao, 
Baoshuo Cai, 
Changqing Zhang, 
Qinghua Hu} \\
\\
College of Intelligence and Computing, Tianjin University, Tianjin, China \\
\texttt{\{caobing, caibaoshuo, zhangchangqing, huqinghua\}@tju.edu.cn}
}

\begin{document}
\maketitle
\input{sec/0_Abstract}    
\input{sec/1_Introduction}

\input{sec/2_RelatedWorks}
\input{sec/3_Method}
\input{sec/4_Experiments}
\input{sec/5_Conclusion}
{
    \small
    \bibliographystyle{ieeenat_fullname}
    \bibliography{main}
}

\newpage
\appendix
\onecolumn
\section{Proof}
\emph{Proof.}
Since $\zeta(\cdot,\,x^*(c))$ is $L$-smooth with respect to its first argument, for any $x,y \in \mathbb{R}^{H \times W \times N}$ we have:
\begin{equation}
\zeta(y,\,x^*(c))
\;\;\le\;\;
\zeta(x,\,x^*(c))
\;+\;
\nabla_{x_t}\zeta(x,\,x^*(c)) \,\cdot\,(y - x)
\;+\;
\frac{L}{2}\,\|y - x\|^2.
\end{equation}
Letting $x = x_t$ and $y = x_{t-1}$ gives a one-step difference inequality:
\begin{equation}
\zeta\bigl(x_{t-1},\,x^*(c)\bigr)
\;-\;
\zeta\bigl(x_t,\,x^*(c)\bigr)
\;\;\le\;\;
\nabla_{x_t}\zeta\bigl(x_t,\,x^*(c)\bigr)\,\cdot\,\bigl(x_{t-1} - x_t\bigr)
\;+\;
\frac{L}{2}\,\|\,x_{t-1} - x_t\,\|^2.
\label{eq:bd}
\end{equation}
In many cases, we are primarily interested in the first-order term (dot product) and regard the second-order term as a manageable constant. Specifically, if we assume$\|x_{t-1} - x_t\|^2 \;\le\;\Delta_t^2$,
so that \eqref{eq:bd} can be relaxed to:
\begin{equation}
\zeta\bigl(x_{t-1},\,x^*(c)\bigr)
\;-\;
\zeta\bigl(x_t,\,x^*(c)\bigr)
\;\;\le\;\;
\nabla_{x_t} \zeta\bigl(x_t,\,x^*(c)\bigr)\cdot \bigl(x_{t-1}-x_t\bigr)
\;+\;
\frac{L}{2}\,\Delta_t^2.
\label{eq:bds}
\end{equation}
Thus, a simple upper bound $\frac{L}{2}\Delta_t^2$ can be carried along in subsequent summations. we decompose:
\begin{align}
x_{t-1} - x_t
\;&=\;
\underbrace{\Bigl(\tfrac{1}{\sqrt{\alpha_t}} - 1\Bigr)\,x_t}_{\text{(I) scaling difference}}
\;+\;
\underbrace{\frac{1}{\sqrt{\alpha_t}}(1 - \alpha_t)\,\nabla_{x_t}\log p(x_t)}_{\text{(II) unconditional gradient}} \notag \\
&+\;
\underbrace{\frac{1}{\sqrt{\alpha_t}}(1 - \alpha_t)\sum_{k=1}^K w_k\,\nabla_{x_t}\log p\bigl(c_k \mid x_t\bigr)}_{\text{(III) multimodal guidance}} \notag \\
&+\;
\underbrace{\sigma_\theta(t)\,z}_{\text{(IV) noise}}.
\end{align}
Plugging this into \eqref{eq:bds}, we have:
\begin{align}
\zeta\bigl(x_{t-1},\,x^*(c)\bigr)
\;-\;
\zeta\bigl(x_t,\,x^*(c)\bigr)
\;&\le\;
\nabla_{x_t} \zeta\bigl(x_t,\,x^*(c)\bigr)\,\cdot\Bigl[\bigl(\tfrac{1}{\sqrt{\alpha_t}} - 1\bigr)\,x_t\Bigr] \nonumber \\
&
+\;
\nabla_{x_t} \zeta\bigl(x_t,\,x^*(c)\bigr)\,\cdot\Bigl[\tfrac{1}{\sqrt{\alpha_t}}(1 - \alpha_t)\,\nabla_{x_t}\,\log p(x_t)\Bigr] \nonumber \\
&
+\;
\nabla_{x_t} \zeta\bigl(x_t,\,x^*(c)\bigr)\,\cdot\Bigl[
\tfrac{1}{\sqrt{\alpha_t}}(1 - \alpha_t)
\sum_{k=1}^K w_k\,\nabla_{x_t}\log p\bigl(c_k \mid x_t\bigr)
\Bigr] \nonumber \\
&
+\;
\nabla_{x_t} \zeta\bigl(x_t,\,x^*(c)\bigr)\,\cdot\bigl[\sigma_\theta(t)\,z\bigr] + \frac{L}{2}\,\Delta_t^2.
\label{eq:bds1}
\end{align}
Leveraging the Markov property of the diffusion model, when $x_t$ is determined, the multi-modal image $c$ is fixed, and the network parameters are constant, the terms independent of $w$ and $x_{t-1}$ do not change. To simplify, we separate these invariants and define:
\begin{align}
G(x_t,c,t)
\;=\;
\nabla_{x_t} \zeta\bigl(x_t,\,x^*(c)\bigr)\,\cdot\Bigl[\bigl(\tfrac{1}{\sqrt{\alpha_t}} - 1\bigr)\,x_t\Bigr]
\;+\;
\nabla_{x_t} \zeta\bigl(x_t,\,x^*(c)\bigr)\,\cdot\Bigl[\tfrac{1}{\sqrt{\alpha_t}}(1 - \alpha_t)\,\nabla_{x_t}\,\log p(x_t)\Bigr].
\end{align}

Then we can rewrite \eqref{eq:bds1} more compactly as:
\begin{align}
\zeta\bigl(x_{t-1},\,x^*(c)\bigr)
\;-\;
\zeta\bigl(x_t,\,x^*(c)\bigr)
\;&\le\;
G(x_t,c,t)
\;+\;
\nabla_{x_t} \zeta\bigl(x_t,\,x^*(c)\bigr)\,\cdot
\Bigl[
\tfrac{1}{\sqrt{\alpha_t}}(1 - \alpha_t)
\sum_{k=1}^K w_k\,\nabla_{x_t}\log p\bigl(c_k \mid x_t\bigr)
\Bigr]\nonumber
\\
&+\;
\nabla_{x_t} \zeta\bigl(x_t,\,x^*(c)\bigr)\,\cdot\bigl[\sigma_\theta(t)\,z\bigr] 
\;+\;
\frac{L}{2}\,\Delta_t^2.
\label{eq:bds2}
\end{align}

Summing from $t=1$ to $T$ in a telescoping manner, we have:
\begin{equation}
\zeta(x_0,\,x^*(c))
\;=\;
\zeta(x_T,\,x^*(c))
\;+\;
\sum_{t=1}^T
\Bigl[
\zeta(x_{t-1},x^*(c))
\;-\;
\zeta(x_t,x^*(c))
\Bigr].
\end{equation}
Applying \eqref{eq:bds2} at each step, we obtain (summing over $t$):

\begin{align}
\zeta\bigl(x_0,\,x^*(c)\bigr)
\;&\le\;
\zeta\bigl(x_T,\,x^*(c)\bigr) \;+\;
\sum_{t=1}^T G\bigl(x_t,c,t\bigr) \;+\;
\sum_{t=1}^T \nabla_{x_t} \zeta\bigl(x_t,\,x^*(c)\bigr)\,\cdot\bigl[\sigma_\theta(t)\,z_t\bigr] \;+\;
\sum_{t=1}^T \tfrac{L}{2}\,\Delta_t^2
\nonumber\\
&+\;
\sum_{t=1}^T
\nabla_{x_t} \zeta\bigl(x_t,\,x^*(c)\bigr)\,\cdot
\Bigl[
\tfrac{1}{\sqrt{\alpha_t}}(1 - \alpha_t)
\sum_{k=1}^K w_k\,\nabla_{x_t}\log p\bigl(c_k \mid x_t\bigr)
\Bigr].
\label{eq:final_telescope}
\end{align}

Finally, recall $x_0 = F(c)$, so $\zeta(x_0,\,x^*(c)) = \zeta\bigl(F(c),\,x^*(c)\bigr)$.  Taking $\mathbb{E}_{c\sim D}$ on both sides of \eqref{eq:final_telescope}, we obtain


\begin{align}
\text{GError}(F) 
&\;=\;
\mathbb{E}_{c \sim D}\bigl[\zeta(x_0,\,x^*(c))\bigr]
\nonumber\\
&\le\;
\mathbb{E}_{c \sim D}\Bigl[\zeta\bigl(x_T,\,x^*(c)\bigr)\Bigr]
\;+\;
\mathbb{E}_{c \sim D}\biggl[\sum_{t=1}^T G\bigl(x_t,c,t\bigr)\biggr]
\;+\;
\mathbb{E}_{c \sim D}\biggl[\sum_{t=1}^T \tfrac{L}{2}\,\Delta_t^2\biggr]
\nonumber\\
&
-\;
\mathbb{E}_{c \sim D}\biggl[
\sum_{t=1}^T
\sum_{k=1}^K 
\tfrac{1}{\sqrt{\alpha_t}}(1 - \alpha_t)w_k\,
\cdot
\Bigl[
\;-\;\nabla_{x_t} \zeta\bigl(x_t,\,x^*(c)\bigr)\,\nabla_{x_t}\log p\bigl(c_k \mid x_t\bigr)\Bigr]
\biggr]
\nonumber\\
&
+\;
\mathbb{E}_{c \sim D}\biggl[
\sum_{t=1}^T 
\nabla_{x_t} \zeta\bigl(x_t,\,x^*(c)\bigr)\,\cdot\bigl[\sigma_\theta(t)\,z_t\bigr]
\biggr]
\nonumber\\
&=\;
\underbrace{\mathbb{E}_{c \sim D}\Bigl[\zeta\bigl(x_T,\,x^*(c)\bigr)\Bigr]
\;+\;
\mathbb{E}_{c \sim D}\biggl[\sum_{t=1}^T G\bigl(x_t,c,t\bigr)\biggr]
\;+\;
\mathbb{E}_{c \sim D}\biggl[\sum_{t=1}^T \tfrac{L}{2}\,\Delta_t^2\biggr]}_{\text{constant}}
\nonumber\\
&
-\;
\sum_{t=1}^T
\Biggl[
\,
\tfrac{1}{\sqrt{\alpha_t}}(1 - \alpha_t)
\underbrace{\sum_{k=1}^K 
\mathbb{E}_{c \sim D}\bigl[w_k\bigr]\,}_{\text{equal to 1}}
\underbrace{\mathbb{E}_{c \sim D}\Bigl[
\;-\;\nabla_{x_t} \zeta\bigl(x_t,\,x^*(c)\bigr)\,\nabla_{x_t}\log p\bigl(c_k \mid x_t\bigr)\Bigr]}_{\text{constant}}\Biggr]
\nonumber\\
&
-\;
\sum_{t=1}^T
\Biggl[
\tfrac{1}{\sqrt{\alpha_t}}(1 - \alpha_t)
\sum_{k=1}^K
\mathrm{Cov}\Bigl(
w_k,\;
-\;\nabla_{x_t} \zeta\bigl(x_t,\,x^*(c)\bigr)\,\nabla_{x_t}\log p\bigl(c_k \mid x_t\bigr)
\Bigr)
\Biggr]
\nonumber\\
&
+\;\sum_{t=1}^T
\Biggl[
\,\mathbb{E}_{c \sim D}\bigl[\nabla_{x_t} \zeta\bigl(x_t,\,x^*(c)\bigr)\bigr]\,
\underbrace{\mathbb{E}_{c \sim D}\bigl[\sigma_\theta(t)\,z_t\bigr]\}}_{\text{equal to 0}}
+\;
\underbrace{\mathrm{Cov}\Bigl(
\nabla_{x_t} \zeta\bigl(x_t,\,x^*(c)\bigr),\;
\sigma_\theta(t)\,z_t
\Bigr)}_{\text{equal to 0}}
\Biggr]
\nonumber\\
&=\;
C \;-\;\sum_{t=1}^T
\Biggl[
\tfrac{1}{\sqrt{\alpha_t}}(1 - \alpha_t)
\sum_{k=1}^K
\mathrm{Cov}\Bigl(
w_k,\;
\underbrace{-\;\nabla_{x_t} \zeta\bigl(x_t,\,x^*(c)\bigr)\,\nabla_{x_t}\log p\bigl(c_k \mid x_t\bigr)}_{\text{alignment Measure}}
\Bigr)
\Biggr]
\end{align}
We revisit the alignment measure
$
-\;\nabla_{x_t}\zeta\bigl(x_t,x^*(c)\bigr)\,\nabla_{x_t}\log p\bigl(c_k \mid x_t\bigr)
$
to elucidate its geometric interpretation. Recall we set
$
v_t
\;=\;
-\;\nabla_{x_t}\zeta\bigl(x_t,x^*(c)\bigr).
$
Intuitively, when $x_t$ is close to the ideal fused image $x^*(c)$, $-\;\nabla_{x_t}\zeta\bigl(x_t,x^*(c)\bigr)$ should point from $x_t$ toward $x^*(c)$. Thus $v_t$ can be seen as a vector indicating the current direction from the generated image to the ideal fused image $x^*(c)$. Let 
$\|v_t\|$ denote the norm of $v_t$. 
Next, let 
$\theta_{t,k}\in[0,\pi]$
be the angle between $v_t$ and $\nabla_{x_t}\log p\bigl(c_k \mid x_t\bigr)$. By definition of the dot product in terms of norms and angles, we have
\begin{equation}
-\;\nabla_{x_t}\zeta\bigl(x_t,x^*(c)\bigr)\,\nabla_{x_t}\log p\bigl(c_k \mid x_t\bigr)
\;=\;
\|v_t\|\,
\bigl\|\nabla_{x_t}\log p\bigl(c_k \mid x_t\bigr)\bigr\|
\cos\bigl(\theta_{t,k}\bigr).
\end{equation}
Because $\|v_t\|$ depends only on $\zeta(\cdot)$ and $x_t$, but not on the specific weight $w_k$ , $\|v_t\|$ can be factored out of the covariance expression. Consequently, in the key covariance term
$
\mathrm{Cov}\Bigl(
\,w_k,\; v_t \,\cdot\,\nabla_{x_t}\log p\bigl(c_k \mid x_t\bigr)
\Bigr),
$
we may treat $\|v_t\|$ as a constant multiplier. Symbolically,
\begin{equation}
\text{GError}(F) 
\;\le\;
C \;-\;\sum_{t=1}^T
\Biggl[
\tfrac{1}{\sqrt{\alpha_t}}(1 - \alpha_t)\|v_t\|\,
\sum_{k=1}^K
\mathrm{Cov}\Bigl(
\,w_k,\;\|\nabla_{x_t}\log p\bigl(c_k \mid x_t\bigr)\|\cos\bigl(\theta_{t,k}\bigr)
\Bigr)
\Biggr]
\end{equation}
As a result, the “directional alignment” with respect to the ideal fusion direction $v_t$ boils down to the projection
$\|\nabla_{x_t}\log p\bigl(c_k \mid x_t\bigr)\|\cos\bigl(\theta_{t,k}\bigr)$. This reveals more transparently that our final bounding in the proof involves a constant factor ($\|v_t\|$) multiplied by the covariance between each weight $w_k$ and the projection of $\nabla_{x_t}\log p\bigl(c_k \mid x_t\bigr)$ onto the direction of $v_t$. Equivalently, one may regard 
$\|\nabla_{x_t}\log p\bigl(c_k \mid x_t\bigr)\|\cos(\theta_{t,k})$
as the “gradient guidance in the ideal fusion direction.” Therefore, the structure of our overall bound becomes more interpretable: the improvement (or reduction) in generalized error stems from how these weights $w_k$ covary with the directional component of the guidance gradient pushing towards $x^*(c)$.

\section{More details about Preliminary}
In DDPM (Denoising Diffusion Probabilistic Models), the forward diffusion process adds noise to a clean sample $x_0$ over multiple steps, eventually transforming it into nearly pure Gaussian noise. This procedure is linear, so one can sample $x_t$ in a single shot at step $t$ via the closed-form expression:
\begin{equation}
x_t = \sqrt{\bar{\alpha}_t} \, x_0 + \sqrt{1 - \bar{\alpha}_t} \, \epsilon, \quad \epsilon \sim \mathcal{N}(0, I),
\label{eq:1}
\end{equation}
where $\alpha_t = 1 - \beta_t$, $\bar{\alpha}_t = \prod_{i=1}^t \alpha_i$, and $\{\beta_t\}_{t=1}^T$ is a predefined variance schedule. As $t$ increases, $x_t$ approaches a nearly pure noise distribution.

To generate a sample during inference, one starts from pure noise $x_T$ and iteratively denoises down to $x_0$. Under a common parameterization, each reverse update step is given by:
\begin{equation}
x_{t-1} = \frac{1}{\sqrt{\alpha_t}} \left( x_t - \frac{1 - \alpha_t}{\sqrt{1 - \bar{\alpha}_t}} \, \epsilon_\theta(x_t, t) \right) + \sigma_\theta(t) \, z,
\label{eq:2}
\end{equation}
where $\sigma_\theta^2(t) = \frac{(1 - \alpha_t)(1 - \bar{\alpha}_{t-1})}{1 - \bar{\alpha}_t}$, $\epsilon_\theta(\cdot)$ is the network’s noise prediction, and $z \sim \mathcal{N}(0, I)$. By iterating from $t = T$ down to $t = 0$, one transforms pure noise into a nearly clean sample.

From the closed-form forward process \cref{eq:1}, If the model $\epsilon_\theta(x_t, t)$ accurately predicts the noise $\epsilon$, one can approximate the “denoised” $\hat{x}_0$ as:
\begin{equation}
\hat{x}_0 = \frac{1}{\sqrt{\bar{\alpha}_t}} \left( x_t - \sqrt{1 - \bar{\alpha}_t} \, \epsilon_\theta(x_t, t) \right).
\label{eq:16}
\end{equation}
This highlights the reverse denoising mechanism: once the correct noise component of $x_t$ is identified, we retrieve a good approximation of the clean data.

From the perspective of stochastic differential equations or variational inference, and based on \cref{eq:1}, the gradient of \( \log p(x_t) \) with respect to \( x_t \) (i.e., the score function) can be expressed as:
\begin{equation}
\nabla_{x_t} \log p(x_t) = -\frac{x_t - \sqrt{\bar{\alpha}_t} x_0}{1 - \bar{\alpha}_t}. 
\label{eq:17}
\end{equation}
Using the estimated \( \hat{x}_0 \) to replace \( x_0 \), and substituting \cref{eq:16},into \cref{eq:17}, we derive:
\begin{equation}
\nabla_{x_t} \log p(x_t) = -\frac{\epsilon_\theta(x_t, t)}{\sqrt{1 - \bar{\alpha}_t}}. 
\label{eq:5}
\end{equation}
In certain applications, such as text-to-image generation and multimodal data fusion, we often wish to incorporate additional conditions during the sampling process. By Bayes' theorem, the gradient of the conditional log-probability with respect to the current sample $x_t$ can be written as
\begin{equation}
\nabla_{x_t} \log p(x_t \mid c)
\;=\; \nabla_{x_t} \log p(x_t) \notag
+\; \nabla_{x_t} \log p(c \mid x_t).
\label{eq:6}
\end{equation}
Here, $c$ represents one or more conditions guiding the generation process.

For $K$ conditions $\{c_k\}_{k=1}^K$, assuming conditional independence given $x_t$, the joint log-probability can be expressed as:
\begin{equation}
\log p(c_1, \ldots, c_K \mid x_t)
\;=\;
\sum_{k=1}^K \log p(c_k \mid x_t).
\label{eq:7}
\end{equation}
In practice, however, these conditions may not be strictly independent, or we may wish to control the influence of each individual condition. Hence, instead of the strict equality in \cref{eq:7}, one frequently uses a weighted approximation:
\begin{equation}
\nabla_{x_t} \log p(c \mid x_t)
\;\approx\;
\sum_{k=1}^K w_k \,\nabla_{x_t} \log p(c_k \mid x_t),
\label{eq:8}
\end{equation}
where $w_k$ is a user-defined weight indicating the relative importance of condition $c_k$. Substituting \cref{eq:8} into \cref{eq:6} then gives:
\begin{align}
\nabla_{x_t} \log p(x_t \mid c)
\;\approx\;
\nabla_{x_t} \log p(x_t) \notag
+\;
\sum_{k=1}^K w_k \,\nabla_{x_t} \log p(c_k \mid x_t).
\label{eq:9}
\end{align}
By adjusting the weights $\{w_k\}$, one can modulate the strength of each condition's contribution to the gradient-based sampling step, thus allowing fine-grained control over the generated samples.

From \cref{eq:2}, \cref{eq:5}, and \cref{eq:9}, we derive the following update equation for the diffusion model:
\begin{equation}
x_{t-1} =
\frac{1}{\sqrt{\alpha_t}} x_t \notag 
+\underbrace{\frac{1}{\sqrt{\alpha_t}} (1 - \alpha_t)\nabla_{x_t} \log p(x_t)}_{\text{Unconditional Guidance}} \notag 
+\underbrace{\frac{1}{\sqrt{\alpha_t}} (1 - \alpha_t)\sum_{k=1}^K w_k \nabla_{x_t} \log p(c_k \mid x_t)}_{\text{Multimodal Guidance}} \notag 
+\underbrace{\sigma_\theta(t) z}_{\text{Noise}}.
\label{eq:10}
\end{equation}
This equation demonstrates that the update step in the diffusion model can be decomposed into three key components: Unconditional Guidance, Multimodal Guidance, and Noise.
This decomposition encourages further exploration of the role of Multimodal Guidance in reducing the model's generalization error and improving conditional generation quality.

\section{Object detection experiment}
\begin{figure*}[t]
  \centering
  \includegraphics[width=1\linewidth]{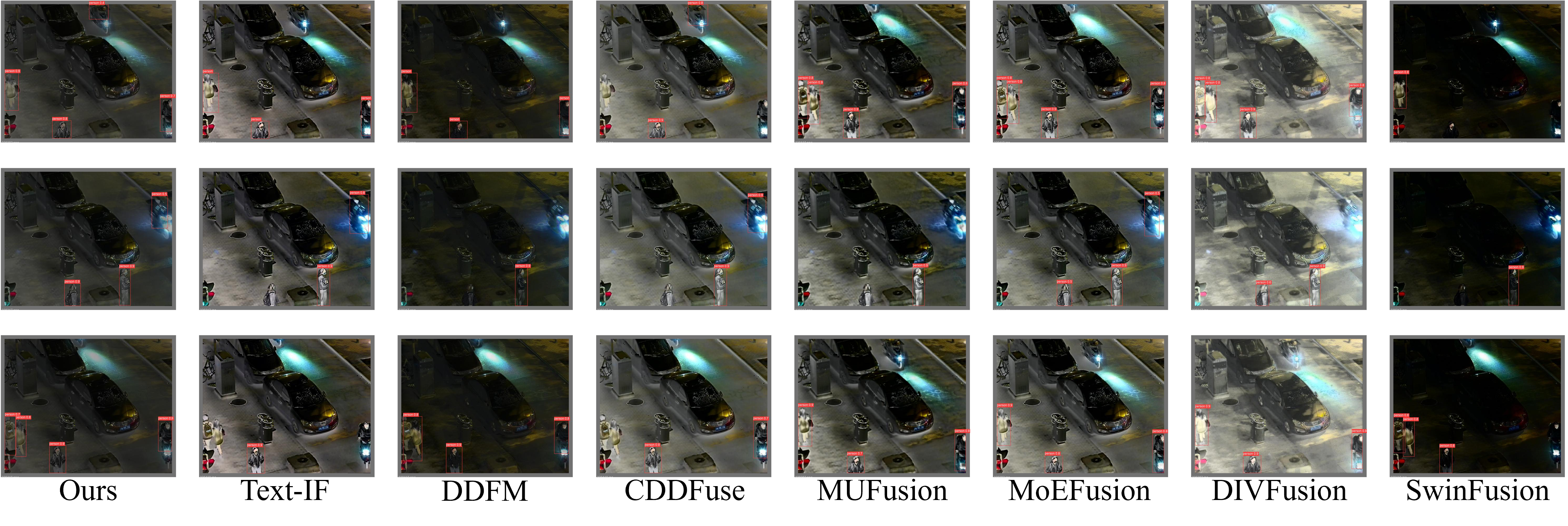}
  \caption{Object detection comparison of our method and the recent proposed competing approaches on LLVIP dataset.}
  \label{fig:100}
\end{figure*}

To evaluate the usability of the fusion results, we use a pre-trained YOLOv5 model to perform pedestrian detection on the LLVIP dataset. The results are shown in ~\ref{fig:100}, demonstrating the usability of the fusion results.

\end{document}

%% file: sec/0_Abstract.tex
\begin{abstract}
Image fusion integrates complementary information from multi-source images to generate more informative results. Recently, the diffusion model, which demonstrates unprecedented generative potential, has been explored in image fusion. However, these approaches typically incorporate predefined multimodal guidance into diffusion, failing to capture the dynamically changing significance of each modality, while lacking theoretical guarantees. To address this issue, we reveal a significant spatio-temporal imbalance in image denoising; specifically, the diffusion model produces dynamic information gains in different image regions with denoising steps. Based on this observation, we Dig into the Diffusion Information Gains (Dig2DIG) and theoretically derive a diffusion-based dynamic image fusion framework that provably reduces the upper bound of the generalization error. Accordingly, we introduce diffusion information gains (DIG) to quantify the information contribution of each modality at different denoising steps, thereby providing dynamic guidance during the fusion process. Extensive experiments on multiple fusion scenarios confirm that our method outperforms existing diffusion-based approaches in terms of both fusion quality and inference efficiency.

\end{abstract}

%% file: sec/1_Introduction.tex
\section{Introduction}
\label{sec:intro}


Image fusion integrates complementary information from various sources to generate informative fused images with high visual quality~\cite{kaur2021image,liang2022fusion,ma2019fusiongan}, thus substantially improving the performance of downstream vision tasks through enhanced scene representations and enriched visual perception. Image fusion can be mainly grouped into three categories: multi-modal image fusion, multi-exposure image fusion, and multi-focus image fusion. 
Multi-modal image fusion (MMF) mainly encompasses Visible-Infrared Image Fusion (VIF) and Medical Image Fusion (MIF) tasks. VIF aims to combine the highlighted thermal targets, especially under extreme conditions, in infrared images and the textural details contributed by visible images~\cite{zhang2020ifcnn,ma2021stdfusionnet}. 
MIF incorporates the active regions of various medical imaging modalities, thereby contributing to diagnostic capabilities~\cite{basu2024systematic}. Different from MMF, MEF~\cite{cao2025test} reconciles the disparity between high- and low-dynamic range images in visual modality, ensuring harmonious lighting appearance, while MFF~\cite{kaur2021image} produces all-in-focus images by blending multiple images captured at different focal depths.

\begin{figure}[t]
  \centering
   \includegraphics[width=1\linewidth]{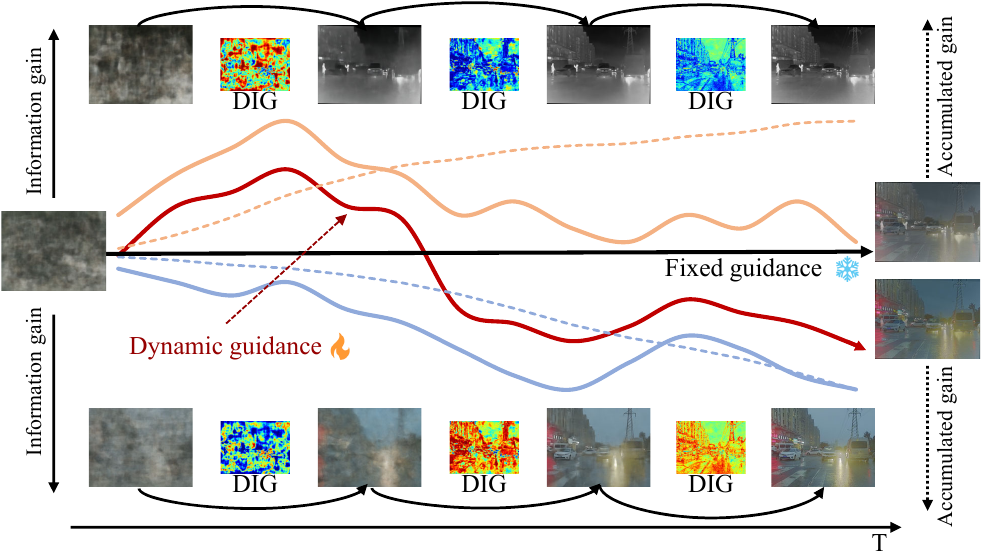}
   \vspace{-2em}
   \caption{Dynamic guidance fusion \textit{vs.} Fixed guidance fusion. The infrared modality, due to its pronounced structural cues, finishes most of its reconstruction earlier, whereas the visible modality, with its abundant texture information, continues to provide significant detail in the later denoising stages.}
   \vspace{-2em}
   \label{fig:1}
\end{figure}

Deep learning-based image fusion techniques, such as CNNs~\cite{amin2019ensemble}, GANs~\cite{rao2023gan}, and Transformers~\cite{ma2022swinfusion}, have outperformed traditional methods~\cite{li1995multisensor,dogra2017multi,yang2009multifocus}; however, their generative capacity usually restricts the detail and realism of the fused images.
Later, diffusion models have emerged as a powerful generative model~\cite{dhariwal2021diffusion}, demonstrating unprecedented potential in image fusion~\cite{zhao2023ddfm}. 
Some works aim to generate high-quality fused images by extracting effective feature representations or incorporating diverse constraints~\cite{cao2025conditional} into diffusion models. 
However, they often employ fixed or predefined multi-modal fusion guidance to the denoising diffusion steps, overlooking the structural dynamism of denoising and failing to produce qualified fusion results in complex scenarios with changing image quality, highlighting the importance of performing dynamic fusion. 

Recently, some studies~\cite{tang2022ydtr} have explored the dynamism in image fusion. For instance, MoE-Fusion~\cite{Cao_2023_ICCV} introduced a dynamic fusion CNN framework with a mixture of experts model, adaptively extracting comprehensive features from diverse modalities. Text-IF~\cite{yi2024text} pioneered the dynamic controllability of image fusion utilizing various text guidance. Furthermore, TTD~\cite{cao2025test} first studied the theoretical foundation of dynamic image fusion during inference. 
Despite their notable empirical performance, these dynamic-oriented fusion methods are mainly limited to CNN-based frameworks, and few works dive into the dynamism of diffusion modeling.
Furthermore, many of these techniques fundamentally rely on heuristic approaches that lack theoretical validation and clear interpretability, leading to unstable fusion results, particularly in complex scenarios.



To address these issues, we reveal the objective of image fusion and dig into diffusion information gains (Dig2DIG) for denoising image fusion with theoretical guarantee. 
Intuitively, image fusion aims to maximize information retention across all modalities~\cite{li2017pixel}. Given that multi-source images jointly determine the fusion result at each step of the diffusion process, the more incremental information of one modality gains at the denoising step contributes more to the overall fusion result, and vice versa.
As illustrated in Figure~\ref{fig:1}, each modality involved in the fusion process demonstrates a distinct denoising pace within the diffusion framework. 
This highlights the dynamic guidance strength of different modalities to effectively preserve and integrate the complementary information offered by each modality.
Building on this insight, we revisit the generalized form of denoising image fusion from the perspective of generalization error, and for the first time prove that the key to enhancing generalization in denoising diffusion fusion lies in the positive correlation of the modality fusion weight and the respective fusion guidance contribution.
Consequently, we derive the Diffusion Information Gains (DIG) as the dynamic fusion weight, which quantifies the contribution of each modality between two noise levels, theoretically enhancing the generalization of the image fusion model, and dynamically highlights the \textit{informative} regions of different sources. 
Extensive experiments on multiple datasets and diverse image fusion tasks demonstrate our superiority in terms of fusion quality and efficiency.
\begin{itemize}
\item 
We theoretically prove that dynamic denoising image fusion outperforms static denoising fusion from the generalization error perspective provably, the key of which lies in the positive covariance between the fusion weight and the respective fusion guidance contribution.
\item 
We introduce Dig2DIG, a simple yet effective dynamic denoising fusion framework. By taking DIG as the dynamic fusion weight, our approach enhances the generalization of the image fusion model while adaptively integrating informative regions from each source.

\item Extensive experiments on diverse fusion tasks validate our superiority. Moreover, an additional exploration of DIG-driven denoising acceleration demonstrates the reasonability of our theory and its potential.

\end{itemize}

%% file: sec/2_RelatedWorks.tex
\section{Related Works}
\label{sec:formatting}
\textbf{Image fusion} aims to integrate complementary information from various sources, such as visible-infrared images, multi-exposure images, and multi-focus images, into a single fused image, thereby improving its visual appearance and downstream task performance. 
Traditional approaches often employ wavelet transforms, multi-scale pyramids, or sparse representations to perform fusion in a transform domain~\cite{wang2005comparative}, while deep learning-based methods (e.g., CNN-, GAN-, or Transformer-based models) learn end-to-end fusion mappings directly in data-driven scheme, which significant enhances the fusion quality compared to traditional methods~\cite{archana2024deep}.
Recently, several fusion approaches based on diffusion models have emerged. For example, DDFM~\cite{zhao2023ddfm} frames the fusion problem as conditional generation within a DDPM framework, utilizing an unconditional pretrained model and expectation-maximization (EM) inference to generate high-quality fused images. CCF~\cite{cao2025conditional} introduces controllable constraints into a pretrained DDPM, allowing the fusion process to adapt to various requirements at each reverse diffusion step, thereby enhancing versatility and controllability. 
Moreover, Text-IF~\cite{yi2024text} incorporates textual semantic guidance into the fusion process, enabling joint image restoration and fusion interactively.
Although some studies have explored dynamic image fusion, the absence of theoretical foundations may yield unstable and unreliable performance in practice, particularly in diffusion models.

\noindent\textbf{Conditional guidance}~\cite{ho2022classifier} in diffusion models typically involves injecting additional priors (such as multi-modal features or textual semantics) at each denoising step, providing a flexible way to steer the final generation or editing outcome. 
Existing studies~\cite{tumanyan2023plug,xu2023infedit} have shown that the guidance on different denoising stages can produce substantially different results, highlighting the importance of dynamic guidance within denoising steps~\cite{cao2025conditional}.
Recently, some dynamic fusion methods were proposed not only for image fusion, but also for more general multi-modal learning. For instance, 
Xue et al.~\cite{xue2023dynamic} employ a Mixture-of-Experts mechanism to integrate multiple experts for multimodal fusion. Han et al.~\cite{han2022trusted} assign the Evidence-driven dynamic weights at the decision level to obtain the trusted fusion decisions, and Zhang et al.~\cite{zhang2023provable} explored the advantages of dynamic fusion and further proposed uncertainty-based fusion weights to enhance the robustness of multimodal learning. 
Although these methods validated the effectiveness of performing dynamic learning, few works reveal the dynamism of conditional guidance in diffusion-based image fusion. 
Most existing methods often assume equal importance for all modalities, overlooking the variations in the information retained by each modality at different denoising stages. This highlights the need for a dynamic guidance mechanism capable of quantifying and utilizing the information gain of each modality.

%% file: sec/3_Method.tex
\begin{figure*}[t]
  \centering
  \includegraphics[width=\linewidth]{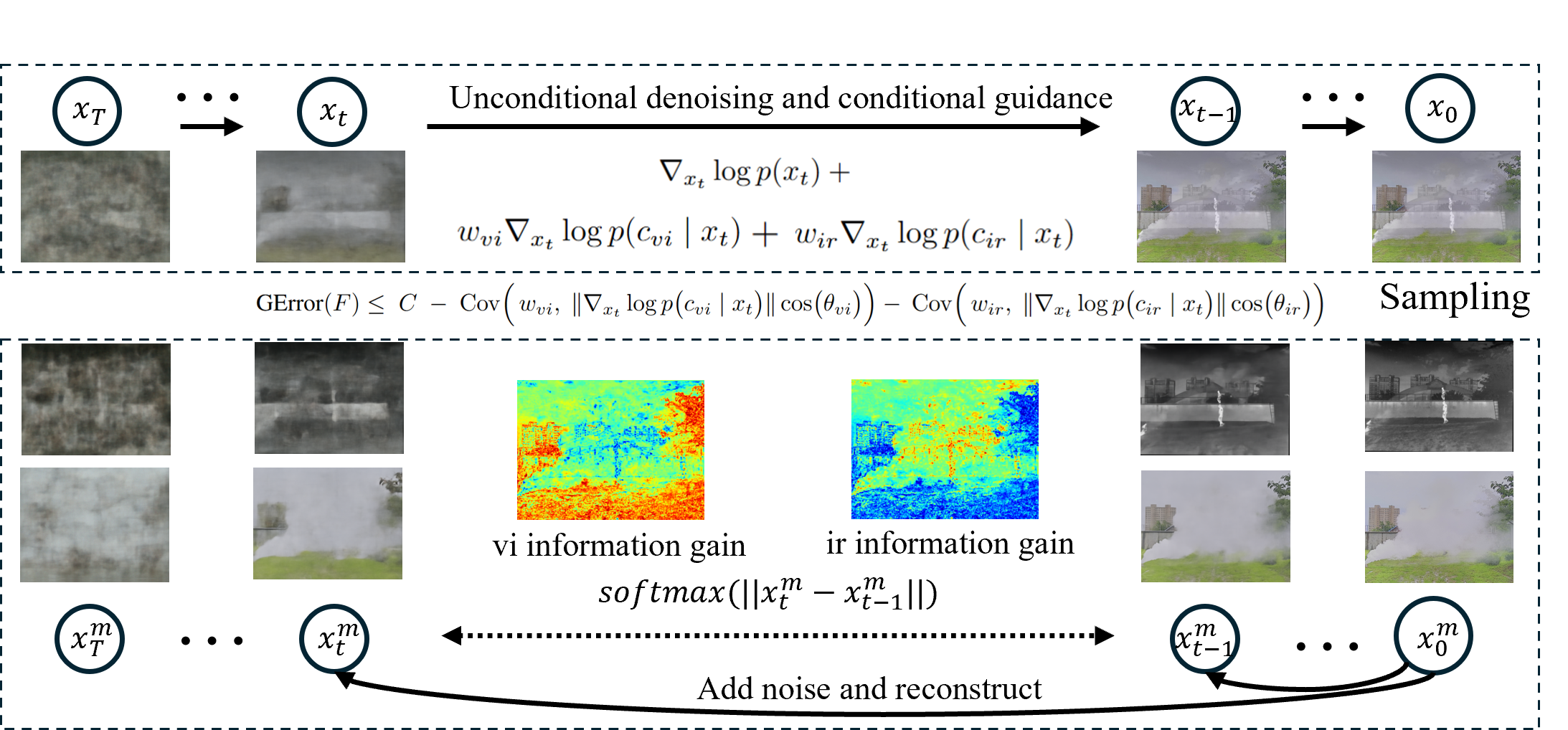}
  \caption{The framework of our Dig2DIG. Deriving from generalization theory, we find that the key to reducing the upper bound of fusion generalization error is ensuring that the projection of the guidance weight and guidance direction onto the ideal fusion direction is positively correlated. To achieve this, we utilize DIG to estimate this projection, providing theoretical guidance for reducing generalization error and effectively incorporating information during the fusion process.}
  \label{fig:2}
\end{figure*}
\section{Method}

In this paper, we dig into the diffusion information gains and propose a denoising-oriented dynamic image fusion framework. We proceed to reveal the Denoising Diffusion Probabilistic Models (DDPM)~\cite{song2020denoising}, the forward diffusion process gradually adds noise to a clean sample $x_0$ until it becomes nearly Gaussian as $
x_t = \sqrt{\bar{\alpha}_t}\, x_0 + \sqrt{1 - \bar{\alpha}_t}\,\epsilon,\quad \epsilon \sim \mathcal{N}(0, I)$, 
where $\alpha_t = 1 - \beta_t$, $\bar{\alpha}_t = \prod_{i=1}^t \alpha_i$, and $\{\beta_t\}$ is a predefined variance schedule.
The characteristic of the diffusion model lies in its ability to estimate the noise of an image during the reverse process.During inference, the noise $x_T$ iteratively denoises via the reverse update:
\begin{equation}
x_{t-1} = \frac{1}{\sqrt{\alpha_t}} \left( x_t - \frac{1 - \alpha_t}{\sqrt{1 - \bar{\alpha}_t}}\, \epsilon_\theta(x_t, t) \right) + \sigma_\theta(t)\, z,
\label{eq:2}
\end{equation}
where $\sigma_\theta^2(t) = (1 - \alpha_t)(1 - \bar{\alpha}_{t-1})/(1 - \bar{\alpha}_t)$, $\epsilon_\theta(\cdot)$ is the network’s noise prediction, and $z \sim \mathcal{N}(0,I)$.

\subsection{Multimodal Guidance}
For the forward process, if $\epsilon_\theta$ accurately reflects the noise in $x_t$, the gradient of $\log p(x_t)$ can be approximated by the score function as $
\nabla_{x_t}\! \log p(x_t) = -\frac{\epsilon_\theta(x_t, t)}{\sqrt{1 - \bar{\alpha}_t}}$. In addition, the conditional guidance $c$ also contributes an additional gradient in many tasks. A practical weighted approximation of the conditional gradient is:
\begin{align}
\nabla_{x_t}\! \log p(x_t \mid c)
&\approx
\nabla_{x_t}\! \log p(x_t)
\notag \\
&\quad+
\sum_{k=1}^K w_k \,\nabla_{x_t}\! \log p(c_k \mid x_t).
\label{eq:9}
\end{align}

Comprehensively, the final update step is given as follows. This succinctly shows how an unconditional term, multimodal guidance, and random noise jointly guide the sample at each denosing step. The full derivations and additional details are presented in Supp.
\begin{align}
x_{t-1}
&= \frac{1}{\sqrt{\alpha_t}} x_t+\underbrace{\sigma_\theta(t)\, z}_{\text{Noise}}
+\underbrace{\frac{1}{\sqrt{\alpha_t}} (1 - \alpha_t)\nabla_{x_t}\! \log p(x_t)}_{\text{Unconditional Guidance}}
\notag \\
& \quad +\underbrace{\frac{1}{\sqrt{\alpha_t}} (1 - \alpha_t)\sum_{k=1}^K w_k \nabla_{x_t}\! \log p(c_k \mid x_t)}_{\text{Multimodal Guidance}}
\label{10}
\end{align}

\subsection{Generalization Error Upper Bound}

Given images $\{c_k\}_{k=1}^{K}$, $c_k \in \mathbb{R}^{H \times W \times N}$ from $K$ sources, the input image combination can be represented as $c = \{c_1, \ldots, c_K\}$.
In the diffusion model, we use $x_t$ to denote the image in the $t$ step of the reverse diffusion process, and the final denoised (fused) result is $x_0 \in \mathbb{R}^{H \times W \times N}$. The overall denoising operator of the diffusion model can be denoted as $F$, i.e., $x_0 = F(c)$.

Let $x^*(c)$ represents the \textit{ideal} fused image conditioned on the multimodal input $c$, and let $\zeta(\cdot)$ be a loss function that measures the discrepancy between a fused image and the ideal image. Assume that $\zeta(\cdot)$ is an $L$-Lipschitz function, under these assumptions, for any unseen data $c \sim D$, we define the Generalization Error as follows:
\begin{equation}
\text{GError}(F) 
= \mathbb{E}_{\,c \sim D} 
\bigl[\zeta\bigl(F(c), x^*(c)\bigr)\bigr].
\end{equation}

Here, $x^*(c)$ denotes the ideal fused image tailored to the input $c$, which reflects the optimal fusion result that we aim to approximate. This expectation quantifies the mean discrepancy between the fused output $F(c)$ and the ideal fused image $x^*(c)$, evaluated on the actual data distribution $D$. A smaller Generalization Error indicates that the model performs better in terms of fusion accuracy on unseen multimodal data.


\begin{theorem}
For a multi-source image fusion operator $F$ that employs diffusion-based conditional guidance, the Generalization Error (GError) can be decomposed as follows:  
(i) A linear combination of projection terms, where each term represents the projection of a single-modal conditional guidance onto the ideal fused modality direction, and  
(ii) A set of constant terms that remain unchanged after model training, given that $\sum_{k=1}^{K} w_{k} = 1$.  
The detailed proof is provided in Supp. A.
\label{thm:MainTheorem}
\vspace{-2em}
\end{theorem}
\begin{equation}
\text{GError}(F) 
\le
C -\sum_{t=1}^T
\Biggl[
A
\sum_{k=1}^K
\mathrm{Cov}\Bigl(
w_k,
B
\Bigr)
\Biggr],
\label{eq:12}
\end{equation}
where $A = \tfrac{1}{\sqrt{\alpha_t}}(1 - \alpha_t) \|v_t\|$ is a timestep-dependent coefficient, and $v_t$ is defined as:  
\begin{equation}
v_t = -\nabla_{x_t} \zeta\bigl(x_t, x^*(c)\bigr),
\end{equation}
which represents the ideal fused modality direction.
\begin{equation}
B = \|\nabla_{x_t}\log p\bigl(c_k \mid x_t\bigr) \|\cos\bigl(\theta_{t,k}\bigr),
\end{equation}
where $\theta_{t,k}$ denotes the angle between $v_t$ and the single-modal conditional guidance direction. Thus, $B$ quantifies the projection of the single-modal conditional guidance onto the ideal fused modality direction.
A larger $B$ indicates that the conditional guidance from modality $k$ is more effective in reducing the discrepancy between the fused image and the ideal fused image $x^*(c)$. To decrease the upper bound of $\text{GError}(F)$, it is necessary to ensure that $\mathrm{Cov}(w_k, B) > 0$. Intuitively, if a particular modality provides stronger guidance in steering the fused image toward the ideal result, its corresponding weight $w_k$ should be higher.
\noindent

In practical applications, the ideal fused image $x^*(c)$ is unobservable and thus the term $B$ defined in \cref{eq:12} cannot be directly computed. Some diffusion-based fusion approaches simplify this problem by assigning the same weight $w_k$ to each modality $c_k$, implying an assumption of equal importance among all modalities. However, multiple studies~\cite{du2023stable,dinh2023rethinking} have observed that the incremental information during the denoising process varies with the and structure and time step. Consequently, different modalities may exhibit different levels of informational contribution across various spatial locations and at different time steps.


Based on this observation, the simplified assumption that the \emph{norm} of the guidance gradient, $\|\nabla_{x_t}\log p(c_k \mid x_t)\|$, does not vary drastically across modalities. Therefore, the main factor affecting $B$ in \cref{eq:12} becomes $\cos(\theta_{t,k})$, such that, the relative sizes of $B$ for different modalities can be approximated by comparing their respective information increments at each reverse diffusion step. According to Theorem~\ref{thm:MainTheorem}, we can use this estimation to adaptively adjust the weights $w_k$ to reduce the overall generalization error.

\subsection{Diffusion Information Gains}
Accordingly, we introduce the concept of Diffusion Information Gains (DIG), which quantifies the extent to which guidance from an individual modality enhances the denoising process at every reverse diffusion step. Specifically, for a single modality $c_k$, let $c_k^t$ denote its noisy degree at timestep $t$, and let $\hat{c}_k^t$ be the corresponding one-step denoised result. The ${DIG}_k(t)$ can be defined as,
\begin{equation}
\mathrm{DIG}_k(t)
\;=\;
l\bigl(\hat{c}_k^t,\;c_k\bigr)
\;-\;
l\bigl(\hat{c}_k^{t-1},\;c_k\bigr),
\label{eq:SMIG_definition}
\end{equation}
where $l(\cdot,\cdot)$ is a function that measures the difference between two images (e.g., an $L^2$ distance). A larger $\mathrm{DIG}_k(t)$ indicates a more substantial reduction of the discrepancy between $\hat{c}_k^t$ and $c_k$ compared to $\hat{c}_k^{t-1}$, suggesting that modality $c_k$ provides more effective guidance at $t$ step.

Following the standard diffusion framework, the noisy image $c_k^t$ at timestep $t$ is generated by:
\begin{equation}
c_k^t 
\;=\; 
\sqrt{\bar{\alpha}_t}\,c_k 
\;+\; 
\sqrt{1-\bar{\alpha}_t}\,\epsilon,
\quad
\epsilon \sim \mathcal{N}(0,\,I),
\end{equation}
where $\bar{\alpha}_t$ controls the noise level. The denoised result $\hat{c}_k^t$ is obtained from $c_k^t$ by the estimated noise:
\begin{equation}
\hat{c}_k^t 
\;=\; 
\frac{1}{\sqrt{\bar{\alpha}_t}}
\Bigl(
c_k^t 
\;-\; 
\sqrt{1-\bar{\alpha}_t}\,\epsilon_\theta\bigl(c_k^t,\,t\bigr)
\Bigr).
\end{equation}


Recalling the upper bound of Generalization Error, the alignment between single-modal guidance and the ``ideal'' fusion direction is characterized by the term $\|\nabla_{x_t}\log p(c_k \mid x_t)\|\cos(\theta_{t,k})$. Intuitively, a larger $\mathrm{DIG}_k(t)$ reflects stronger alignment with the ideal direction, because it indicates that modality $c_k$ is contributing more effectively to reducing the discrepancy between the current estimate and its clean target. Therefore, $\mathrm{DIG}_k(t)$ can be viewed as a practical proxy for the alignment measure discussed earlier.

\subsection{Dynamic Fusion with DIG}
Given $\mathrm{DIG}_k(t)$ for each modality $c_k$, we propose to dynamically weight the guidance contributions to the fused image based on their diffusion information gains. At each denosing step $t$, the weights $\{w_k\}$ is computed by normalizing the DIG values across the modalities (e.g., via a softmax function):
\begin{equation}
w_k(t)
\;=\;
\frac{\exp\!\bigl(\mathrm{DIG}_k(t)\bigr)}
{\sum_{j=1}^{K}\exp\!\bigl(\mathrm{DIG}_j(t)\bigr)}.
\label{eq:SMIG_softmax}
\end{equation}
By incorporating DIG-based weights, the fused result more accurately reflects the relative contributions of each modality at each timestep, ultimately leading to a lower fusion error and better generalization performance.

%% file: sec/4_Experiments.tex
\section{Experiments}

\begin{figure*}[t]
  \centering
  \includegraphics[width=\linewidth]{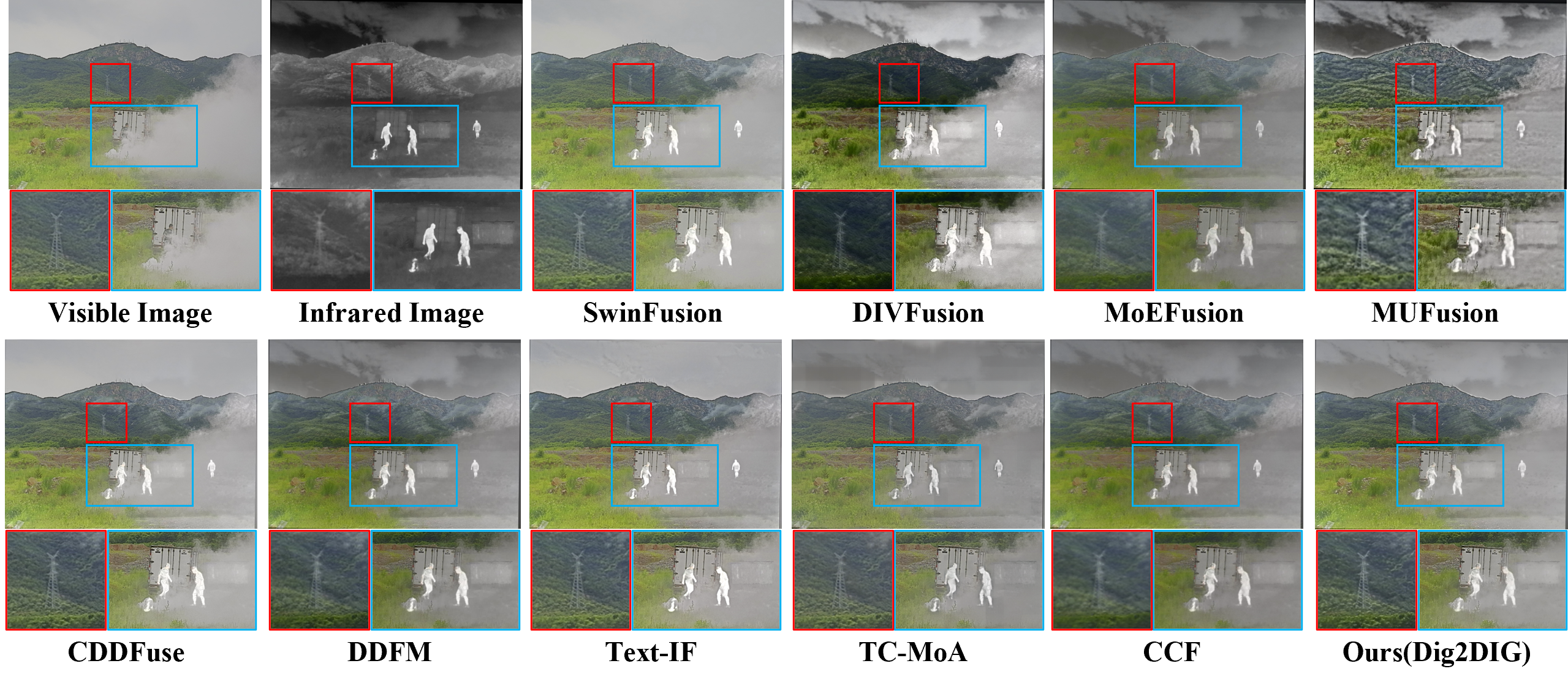}
  \vspace{-2.5em}
  \caption{ Qualitative comparisons of our method and the competing approaches on M3FD Dataset.}
  \label{fig:3}
\end{figure*}

\begin{table*}[t]
\caption{Quantitative comparisons on the LLVIP, M3FD, and MSRS datasets. The best, second best, and third best results are highlighted in \textcolor{Red}{red}, \textcolor{Blue}{blue}, and \textcolor{Green}{green}, respectively.}
\vspace{-0.5em}
    \centering
    \fontsize{12}{15}\selectfont\setlength{\tabcolsep}{1mm}
    \resizebox{\textwidth}{!}{
    \begin{tabular}{l | c c c c c c | c c c c c c | c c c c c c}
        \toprule
        & \multicolumn{6}{c|}{\textbf{LLVIP Dataset}} & \multicolumn{6}{c|}{\textbf{M3FD Dataset}} & \multicolumn{6}{c}{\textbf{MSRS Dataset}} \\ 
        \cmidrule(lr){2-7} \cmidrule(lr){8-13} \cmidrule(lr){14-19}
        \textbf{Method} 
        & \textbf{PSNR$\uparrow$} & \textbf{SSIM$\uparrow$} & \textbf{MSE$\downarrow$} & \textbf{Nabf$\downarrow$} & \textbf{CC$\uparrow$} & \textbf{LPIPS$\downarrow$} 
        & \textbf{PSNR$\uparrow$} & \textbf{SSIM$\uparrow$} & \textbf{MSE$\downarrow$} & \textbf{Nabf$\downarrow$} & \textbf{CC$\uparrow$} & \textbf{LPIPS$\downarrow$}
        & \textbf{PSNR$\uparrow$} & \textbf{SSIM$\uparrow$} & \textbf{MSE$\downarrow$} & \textbf{Nabf$\downarrow$} & \textbf{CC$\uparrow$} & \textbf{LPIPS$\downarrow$}\\
        \midrule
        SwinFusion 
        & 32.33 & 0.81 & 2845 & 0.023 & 0.67 & 0.321
        & 31.73 & \textcolor{Blue}{\textbf{1.40}} & 3853 & 0.021 & 0.51 & 0.289
        & \textcolor{Blue}{\textbf{39.34}} & \textcolor{Blue}{\textbf{1.41}} & 1755 & \textcolor{Blue}{\textbf{0.002}} & 0.59 & 0.298 \\
        
        DIVFusion  
        & 21.60 & 0.82 & 6450 & 0.044 & 0.66 & 0.350
        & 26.19 & 1.20 & 4099 & 0.083 & 0.51 & 0.377
        & 18.49 & 0.69 & 10054 & 0.100 & 0.52 & 0.462 \\
        
        MoE-Fusion  
        & 31.70 & 1.12 & 2402 & 0.034 & \textcolor{Green}{\textbf{0.69}} & 0.324
        & \textcolor{Blue}{\textbf{33.15}} & 1.37 & 3462 & 0.012 & 0.47 & 0.303
        & 38.21 & 1.35 & 2637 & 0.030 & 0.60 & 0.298 \\
        
        MUFusion   
        & 31.64 & 1.10 & 2069 & 0.030 & 0.65 & 0.320
        & 29.82 & 1.29 & 2733 & 0.071 & 0.50 & 0.349
        & 36.02 & 1.25 & 1701 & 0.037 & 0.037 & 0.370 \\
        
        CDDFuse    
        & 32.13 & 1.18 & 2545 & 0.016 & 0.67 & 0.335
        & 31.75 & \textcolor{Green}{\textbf{1.40}} & 3715 & 0.030 & 0.52 & \textcolor{Blue}{\textbf{0.278}}
        & 37.76 & 1.30 & 2485 & 0.022 & 0.59 & 0.335 \\
        
        DDFM       
        & \textcolor{Red}{\textbf{36.10}} & 1.18 & \textcolor{Green}{\textbf{2056}} & \textcolor{Blue}{\textbf{0.004}} & 0.67 & \textcolor{Blue}{\textbf{0.310}}
        & 30.87 & 1.40 & \textcolor{Blue}{\textbf{2221}} & \textcolor{Red}{\textbf{0.007}} & \textcolor{Green}{\textbf{0.56}} & 0.303
        & 38.19 & 1.39 & \textcolor{Blue}{\textbf{1367}} & \textcolor{Green}{\textbf{0.004}} & \textcolor{Red}{\textbf{0.66}} & \textcolor{Blue}{\textbf{0.287}} \\
        
        Text-IF    
        & 31.22 & 1.18 & 2460 & 0.031 & 0.69 & \textcolor{Green}{\textbf{0.312}}
        & \textcolor{Red}{\textbf{34.01}} & 1.39 & 3470 & 0.037 & 0.48 & \textcolor{Red}{\textbf{0.277}}
        & \textcolor{Red}{\textbf{41.93}} & 1.37 & 2494 & 0.027 & 0.60 & 0.298 \\
        
        TC-MoA     
        & 33.00 & \textcolor{Green}{\textbf{1.20}} & 2790 & 0.017 & 0.67 & 0.332
        & 31.07 & 1.40 & 2516 & 0.011 & 0.53 & 0.289
        & 37.73 & \textcolor{Green}{\textbf{1.40}} & 1640 & 0.005 & 0.62 & \textcolor{Green}{\textbf{0.293}} \\
        
        CCF        
        & \textcolor{Green}{\textbf{33.12}} & \textcolor{Blue}{\textbf{1.22}} & \textcolor{Blue}{\textbf{1658}} & \textcolor{Green}{\textbf{0.006}} & \textcolor{Blue}{\textbf{0.70}} & 0.334
        & 31.51 & 1.40 & \textcolor{Green}{\textbf{2271}} & \textcolor{Green}{\textbf{0.010}} & \textcolor{Blue}{\textbf{0.56}} & 0.291
        & 38.00 & 1.38 & \textcolor{Green}{\textbf{1410}} & 0.006 & \textcolor{Blue}{\textbf{0.64}} & 0.319 \\
        
        Dig2DIG (ours) 
        & \textcolor{Blue}{\textbf{33.74}} & \textcolor{Red}{\textbf{1.23}} & \textcolor{Red}{\textbf{1464}} & \textcolor{Red}{\textbf{0.001}} & \textcolor{Red}{\textbf{0.73}} & \textcolor{Red}{\textbf{0.298}}
        & \textcolor{Green}{\textbf{31.83}} & \textcolor{Red}{\textbf{1.41}} & \textcolor{Red}{\textbf{2216}} & \textcolor{Blue}{\textbf{0.009}} & \textcolor{Red}{\textbf{0.57}} & \textcolor{Green}{\textbf{0.287}}
        & \textcolor{Green}{\textbf{39.07}} & \textcolor{Red}{\textbf{1.42}} & \textcolor{Red}{\textbf{1366}} & \textcolor{Red}{\textbf{0.001}} & \textcolor{Green}{\textbf{0.63}} & \textcolor{Red}{\textbf{0.282}} \\
        \bottomrule
    \end{tabular}
    }
    \label{tab:llvip-m3fd-msrs}
\end{table*}

\subsection{Experimental Setting}
\textbf{Datasets.} In our experiments, we evaluate the proposed method on three key image fusion tasks: 
Visible-Infrared Image Fusion (VIF), Multi-Focus Fusion (MFF), and 
Multi-Exposure Fusion (MEF).
For VIF, we use the LLVIP~\cite{jia2021llvip}, M3FD~\cite{liu2022target}, and MSRS~\cite{Tang2022PIAFusion} datasets, each providing paired visible and infrared images under a variety of scenarios. In the MFF task, we adopt the MFFW dataset~\cite{zhang2021benchmarking} to merge images that focus on different regions into a single, fully focused output. For the MEF task, we employ the MEFB dataset ~\cite{zhang2021deep} to assess the performance of combining images captured at various exposure levels.

\noindent\textbf{Implementation Details.} Our approach is built upon a pre-trained diffusion model~\cite{dhariwal2021diffusion}, and crucially, it does not require any additional training or fine-tuning. We leverage the pretrained network directly for each fusion task, thereby eliminating the need for task-specific supervision. All experiments were conducted on NVIDIA RTX A6000 GPUs.

\noindent\textbf{Evaluation Metrics.} We evaluate fusion quality using both qualitative and quantitative approaches. Qualitative assessment relies on subjective visual inspection, focusing on clear textures and natural color representation. For the Visible-Infrared Image Fusion (VIF) task, we use Peak Signal-to-Noise Ratio (PSNR), Structural Similarity (SSIM), Mean Squared Error (MSE), Noise Amplification (Nabf), Correlation Coefficient (CC), and Learned Perceptual Image Patch Similarity (LPIPS). For the MFF and MEF tasks, we employ Standard Deviation (SD), Edge Intensity (EI), Entropy (EN), Average Gradient (AG), Spatial Frequency (SF), and Mutual Information (MI).

\subsection{Comparison on Visible-Infrared Image Fusion}
For Visible-Infrared image fusion, we compare our method with the state-of-the-art methods: SwinFusion~\cite{ma2022swinfusion}, DIVFusion~\cite{Tang2022DIVFusion}, 
MOEFusion~\cite{Cao_2023_ICCV},
MUFusion~\cite{cheng2023mufusion}, CDDFuse~\cite{zhao2023cddfuse}, DDFM~\cite{zhao2023ddfm}, Text-IF~\cite{yi2024text}, TC-MoA~\cite{zhu2024task}, and CCF~\cite{cao2025conditional}.

\noindent\textbf{Quantitative Comparisons.}
Table~\ref{tab:llvip-m3fd-msrs} presents the quantitative results on three infrared-visible datasets (LLVIP, M3FD, and MSRS) under six evaluation metrics. Our proposed method (Dig2DIG) achieves leading performance on the majority of these metrics without requiring any training procedure.
On the \textbf{LLVIP} dataset, Dig2DIG attains the best SSIM, MSE, CC, and LPIPS scores, while also showing a notably low Nabf. For instance, our MSE (1464) not only outperforms the second-best value (1658) but is also indicative of improved fidelity to the original images. Additionally, our SSIM (1.23) surpasses previous methods, demonstrating superior structural preservation.
In the \textbf{M3FD} dataset, our method again secures top rankings in several metrics, including SSIM and CC. The reduction of MSE from 2221 (second-best) to 2216 underlines our consistent fidelity benefits, while the improvements in SSIM highlight enhanced structural similarity.
Meanwhile, on the \textbf{MSRS} dataset, Dig2DIG achieves the best SSIM, MSE, and LPIPS scores. The lower MSE (1366) suggests stronger detail retention, and the improved LPIPS (0.282) indicates better perceptual quality. Our results on these datasets confirm that incorporating diffusion information gains (DIG) effectively captures and balances the contributions from each modality during denoising, offering robust fusion without additional fine-tuning.

\noindent\textbf{Qualitative Comparisons.} By leveraging Diffusion Information Gains, our method effectively captures high-information modalities during the reverse diffusion process, resulting in more accurate and robust image fusion. As illustrated in Fig.~\ref{fig:3}, the red boxes highlight regions where Dig2DIG yields sharper structural details, whereas competing methods such as CCF and MOFfusion exhibit blur and lose fine-grained information. Furthermore, the blue boxes in Fig.~\ref{fig:3} demonstrate our method’s superior ability to incorporate infrared cues while retaining critical visual content, surpassing methods like TC-MoA and Text-IF. These observations underscore the advantages conferred by DIG, which help preserve both background clarity and crucial texture information.

\subsection{Evaluation on Multi-Focus Fusion}
\begin{figure*}[t]
  \centering
  \includegraphics[width=\linewidth]{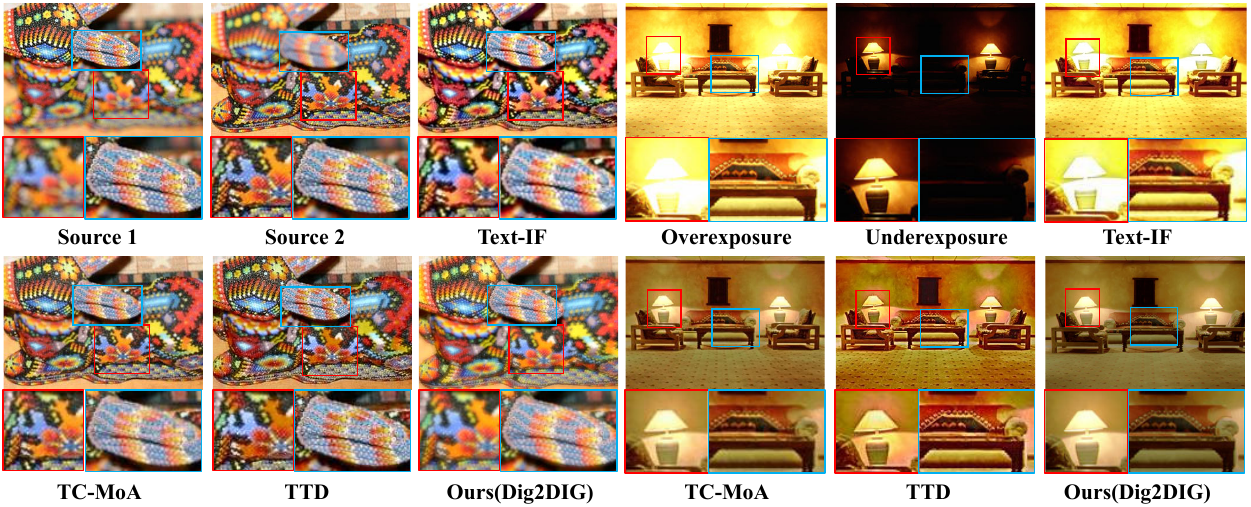}
  \vspace{-2em}
  \caption{Qualitative comparisons of our method and the competing approaches on MFFW Dataset and MEFB Dataset}
  \label{fig:6}
\end{figure*}
\begin{table*}[t]
\caption{Performance comparison on the MFFW   Dataset and the MEFB  Dataset. The best, runner-up, and third best results are highlighted.}
\vspace{-0.8em}
\fontsize{6}{7}\selectfont\setlength{\tabcolsep}{1.8mm}
\centering
\resizebox{\textwidth}{!}{%
\begin{tabular}{l| c c c c c c | c c c c c c}
\toprule
 & \multicolumn{6}{c|}{\textbf{MFFW Dataset}} & \multicolumn{6}{c}{\textbf{MEFB Dataset}} \\
\cmidrule(lr){2-7} \cmidrule(lr){8-13}
\textbf{Method} & \textbf{SD$\uparrow$} & \textbf{EI$\uparrow$} & \textbf{EN$\uparrow$} & \textbf{AG$\uparrow$} & \textbf{SF$\uparrow$} & \textbf{MI$\uparrow$} & \textbf{SD$\uparrow$} & \textbf{EI$\uparrow$} & \textbf{EN$\uparrow$} & \textbf{AG$\uparrow$} & \textbf{SF$\uparrow$} & \textbf{MI$\uparrow$} \\
\midrule
FusionDN & \textcolor{Green}{\textbf{66.59}} & \textcolor{Red}{\textbf{17.20}}   & \textcolor{Green}{\textbf{7.45}}  & \textcolor{Blue}{\textbf{6.74}}  & \textcolor{Blue}{\textbf{22.27}} & 3.37   & 61.50 & \textcolor{Green}{\textbf{19.55}} & 7.29   & 7.56   & 21.05   & 3.47 \\
U2Fusion & 64.88                        & 11.97                         & 6.93                         & 5.56                         & 18.74                         & 3.251  & \textcolor{Blue}{\textbf{67.83}}  & 19.54                         & \textcolor{Green}{\textbf{7.37}}  & \textcolor{Blue}{\textbf{8.08}}  & \textcolor{Green}{\textbf{22.19}} & 3.38 \\
DeFusion & 52.75                        & 10.60                          & 6.80                          & 4.32                         & 14.12                         & 2.92   & 54.75  & 12.55                         & 7.28   & 4.76   & 12.72   & 3.89 \\
DDFM     & \textcolor{Blue}{\textbf{67.30}}  & 14.32                         & \textcolor{Blue}{\textbf{7.51}}  & 3.82                         & 13.40                          & \textcolor{Blue}{\textbf{5.71}}  & 56.34  & 11.95                         & 7.30    & 4.47   & 12.21   & \textcolor{Red}{\textbf{8.49}} \\
Text-IF  & 62.51                        & 12.73                         & 6.39                         & 4.82                         & 17.26                         & 3.41   & \textcolor{Green}{\textbf{66.27}}  & \textcolor{Blue}{\textbf{20.01}} & 7.37   & \textcolor{Green}{\textbf{7.72}}  & 21.58   & 3.30 \\
TC-MoA  & 50.27                        & 12.18                         & 7.07                         & 4.82                         & 15.64                         & 3.39   & 57.55  & 17.65                         & 7.35   & 6.95   & 20.67   & \textcolor{Green}{\textbf{4.45}} \\
TTD      & 52.86                        & \textcolor{Green}{\textbf{15.94}} & 7.10                          & \textcolor{Green}{\textbf{6.38}} & \textcolor{Green}{\textbf{21.99}} & \textcolor{Green}{\textbf{4.54}}  & 54.22  & 19.10                          & \textcolor{Red}{\textbf{7.39}}  & 7.70    & \textcolor{Blue}{\textbf{23.51}} & 3.59 \\
Dig2DIG (ours)     & \textcolor{Red}{$\textbf{72.95}$}  & \textcolor{Blue}{\textbf{16.64}} & \textcolor{Red}{\textbf{7.87}}  & \textcolor{Red}{\textbf{6.75}}  & \textcolor{Red}{\textbf{22.60}}   & \textcolor{Red}{\textbf{5.97}}  & \textcolor{Red}{\textbf{75.05}}  & \textcolor{Red}{\textbf{20.21}} & \textcolor{Blue}{\textbf{7.38}} & \textcolor{Red}{\textbf{8.10}}  & \textcolor{Red}{\textbf{23.60}}  & \textcolor{Blue}{\textbf{6.87}} \\
\bottomrule
\end{tabular}%
}
\label{tab:combined}
\end{table*}
For multi-focus image fusion, we compare our method with the state-of-the-art methods: FusionDN~\cite{xu2020aaai}, U2Fusion~\cite{xu2020u2fusion}, 
DeFusion~\cite{liang2022fusion},
DDFM~\cite{zhao2023ddfm}, Text-IF~\cite{yi2024text}, TC-MoA~\cite{zhu2024task}, and TTD~\cite{cao2025test}.

\noindent\textbf{Quantitative Comparisons.} 
We evaluate our approach on the MFFW dataset using six metrics (SD, EI, EN, AG, SF, and MI). As shown in Table~2 (left), Dig2DIG outperforms competing methods on five of these six indicators by notable margins. In particular, our method achieves the highest SD (72.95), which is 5.65 above the second-best (67.30), reflecting enhanced contrast and clarity. We also secure top positions in EN (7.87), AG (6.75), SF (22.60), and MI (5.97), suggesting superior retention of details and overall information. Although FusionDN slightly outperforms Dig2DIG in EI, our model still ranks second. These results validate the efficacy of our dynamic diffusion-based fusion framework in handling multi-focus imagery. This result demonstrates the effectiveness of our method.

\noindent\textbf{Qualitative Comparisons.} In the multi-focus MFFW dataset, our approach continues to exhibit robust performance in preserving fine-grained details and color fidelity. As shown in Fig.~\ref{fig:6}, the blue box highlights how our method captures subtle textures and accurately reflects the original hues, even under challenging multi-focus conditions. In contrast, other methods such as TTD and Text-IF struggle to maintain the same level of clarity or color consistency. These results underscore the effectiveness of our framework in handling multi-focus scenes while faithfully retaining both structural and color information.the improved visual quality achieved by our method further demonstrates its superiority in complex fusion scenarios.

\subsection{Evaluation on Multi-Exposure Fusion}
For multi-exposure image fusion, we compare our method with the state-of-the-art methods: FusionDN~\cite{xu2020aaai}, U2Fusion~\cite{xu2020u2fusion}, 
DeFusion~\cite{liang2022fusion},
DDFM~\cite{zhao2023ddfm}, Text-IF~\cite{yi2024text}, TC-MoA~\cite{zhu2024task}, and TTD~\cite{cao2025test}.

\noindent\textbf{Quantitative Comparisons.}As shown in Table~2 (right), we evaluate our method on the MEFB dataset using SD, EI, EN, AG, SF, and MI. Dig2DIG obtains the best performance on four of these metrics (SD, EI, AG, SF), with values of 75.05, 20.21, 8.10, and 23.60, respectively. While TTD achieves a slightly higher EN (7.39 vs.\ 7.38) and DDFM outperforms us in MI (8.49 vs.\ 6.87), our method still ranks second in both metrics. These results confirm the robustness of our framework in multi-exposure scenarios, demonstrating its effectiveness in highlighting important regions, preserving image details, and combining the different exposures cohesively.

\noindent\textbf{Qualitative Comparisons.}
On the multi-exposure dataset, Text-IF frequently suffers from overexposure, causing loss of crucial details in high-luminance regions. In contrast, our method, Dig2DIG, effectively synthesizes information across different exposure levels, preserving both brightness and texture details. By maintaining an ideal balance of saturation and clarity, Dig2DIG outperforms competing approaches in overall visual fidelity and detail retention.

\subsection{Discussion}
\begin{figure}[t]
  \centering
  \includegraphics[width=\linewidth]{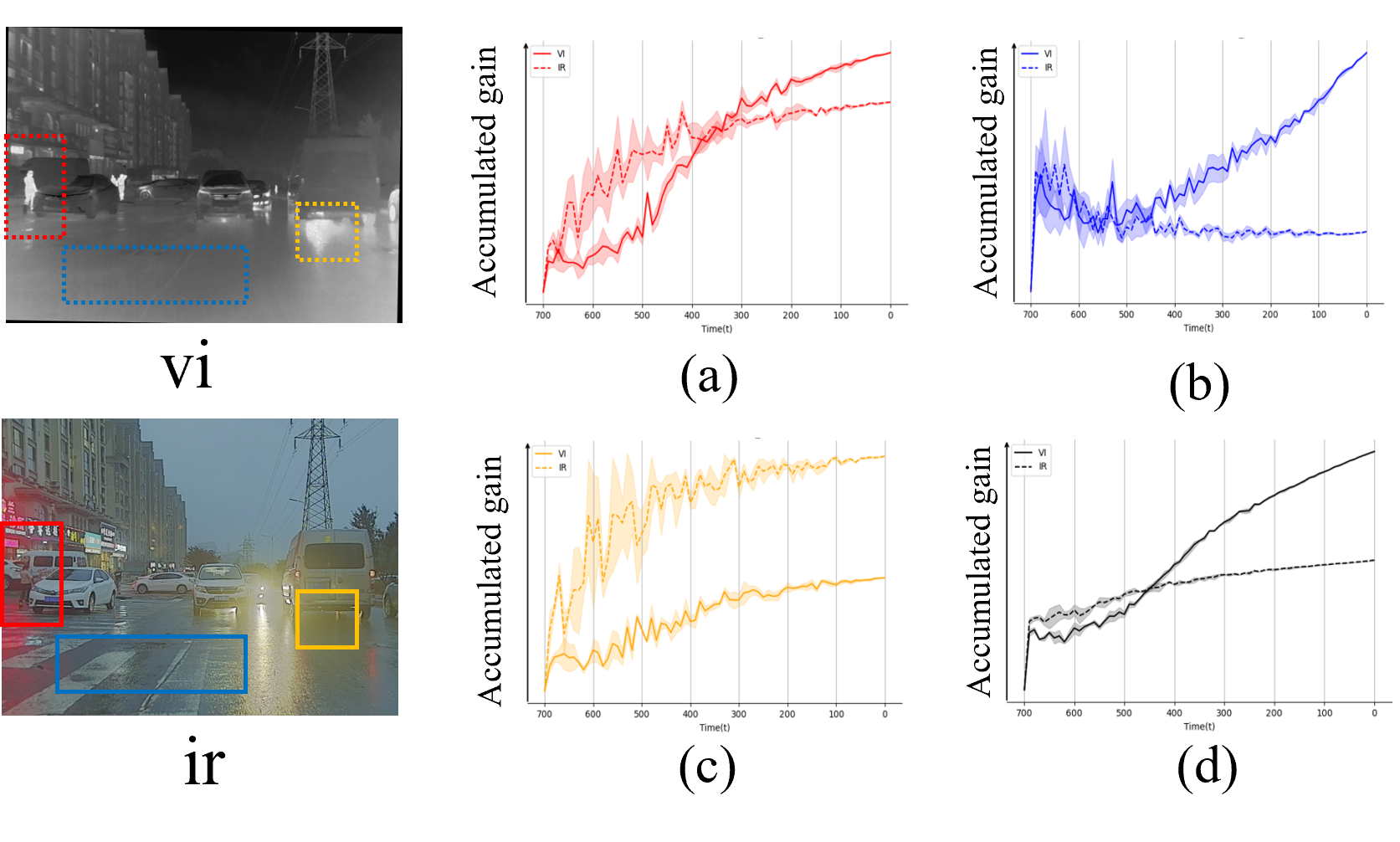}
  \vspace{-2em}
  \caption{The spatio-temporal imbalance of information gains.}
  \vspace{-2em}
  \label{fig:d1}
\end{figure}
\paragraph{Discussion of spatio-temporal imbalance.}
Figure~\ref{fig:d1} presents the spatio-temporal imbalance of information gains on an infrared-visible image pair. The curves in each plot indicate the mean and variance of the cumulative Diffusion Information Gains starting from a certain noise level. Figure(d) depicts the global cumulative DIG for the entire image, revealing that the restoration speeds of different modalities vary during the inverse denoising process of the diffusion model. Meanwhile, Figures~\ref{fig:d1}(a), \ref{fig:d1}(b), and \ref{fig:d1}(c) illustrate the cumulative DIG of the two modalities within different regions of the image.
In the region analyzed by Figure~\ref{fig:d1}(a), the infrared modality exhibits a clearly discernible structure, leading to a larger information gain when noise is relatively high. In contrast, the region shown in Figure~\ref{fig:d1}(b) has almost no meaningful information in the infrared modality, resulting in only a small change in its cumulative DIG. As for the region in Figure~\ref{fig:d1}(c), since the visible modality contains a limited amount of information, the cumulative DIG of the infrared modality surpasses that of the visible modality. These observations indicate that, throughout the denoising process, not only do information discrepancies exist between different modalities, but such spatio-temporal imbalance of information gains also persists across various regions of the image.
\begin{table}[t]
    \caption{Comparison of different distance measures on the M3FD dataset. 
    $\varnothing$ denotes the baseline without any distance measure.}
    \vspace{-0.8em}
\fontsize{9}{10}\selectfont\setlength{\tabcolsep}{4mm}
    \centering
    \resizebox{1\columnwidth}{!}{
    \begin{tabular}{lcccc}
    \toprule
    \textbf{Metric} & \textbf{PSNR}~$\uparrow$ & \textbf{MSE}~$\downarrow$ & \textbf{CC}~$\uparrow$ & \textbf{LPIPS}~$\downarrow$ \\
    \midrule
    $\varnothing$ & 30.87 & 2221 & 0.568 & 0.303 \\
    $\ell_1$ & 31.25 & 2220 & 0.569 & 0.293 \\
    SSIM & 31.69 & 2245 & 0.571 & 0.297 \\
    $\ell_2$ & \textbf{31.83} & \textbf{2216} & \textbf{0.573} & \textbf{0.287} \\
    \bottomrule
    \end{tabular}
    }
    \label{tab:m3fd_l_choice}
    \vspace{-0.5em}
\end{table}

\noindent\textbf{Discussion of Efficiency.}
To reduce the overhead of computing DIG  at each reverse diffusion step, we introduce a hyperparameter $S$ that specifies the interval at which DIG is calculated. In other words, instead of computing DIG at every step, it is updated every $S$ steps by computing the difference over $S$ steps rather than between consecutive steps. $S=1$ means DIG is computed at every step, and $S=10$ means it is computed once per ten denoising steps.
\begin{table}[t]
\centering
\caption{Performance of different DIG intervals $S$ on M3FD.}
\vspace{-0.8em}
\fontsize{9}{10}\selectfont\setlength{\tabcolsep}{4mm}
\begin{tabular}{c|cccc}
\toprule
\textbf{$S$} & \textbf{SSIM}~$\uparrow$ & \textbf{MSE}~$\downarrow$ & \textbf{CC}~$\uparrow$ & \textbf{LPIPS}~$\downarrow$ \\
\midrule
1  & 1.35 & 2562 & 0.521  & 0.3082 \\
5  & 1.38 & 2321 & 0.533  & 0.2935 \\
10 & \textbf{1.41} & \textbf{2215} & \textbf{0.5728} & \textbf{0.2870} \\
20 & 1.40 & 2220 & 0.5698 & 0.2890 \\
\bottomrule
\end{tabular}
\label{tab:dig_frequency}
\vspace{-2em}
\end{table}

As shown in Table~\ref{tab:dig_frequency}, using $S=10$ achieves the best balance between fusion quality and computational cost. When $S=1$ or $S=5$, the difference between adjacent steps is too small, making it difficult to accurately capture information gains, which slightly affects the performance. On the other hand, when $S=20$, the performance degrades marginally, presumably because DIG is updated too infrequently to capture finer-grained changes in the dynamic guidance. Therefore, we adopt $S=10$ to reduce computational overhead while preserving high-quality fusion results.
``DIG-$N$'' denotes our method with a total of $N$ reverse diffusion steps.Based on the results in Table~\ref{tab:dig_total_steps}, increasing the total number of reverse diffusion steps generally improves performance but also significantly increases runtime. We find that ``DIG-25'' effectively strikes a balance between runtime and fusion quality: it offers favorable performance while keeping the average runtime to 52 seconds, which is significantly faster than CCF’s 633 seconds and moderately better than DDFM’s 180 seconds. 

Note that in the early stages of the reverse diffusion process, the noise level is high and the variance of DIG is large, which often makes the information gain inaccurate or ineffective. Based on this , and in order to fuse information more efficiently, Dig2DIG employs larger denoising steps at higher noise levels and smaller denoising steps at lower noise levels. This approach ensures that, when the noise is sufficiently reduced, the valuable features of each modality can be more deeply integrated, thus effectively leveraging the higher information gain available at lower noise levels and achieving more efficient information fusion.

\noindent\textbf{Discussion of the choice of $l$.} To determine a suitable function for computing \( l \), we conduct experiments on the M3FD dataset using different metric functions, including \( \ell_1 \), SSIM, and \( \ell_2 \), while considering the case without any metric function as the "baseline," denoted by $\varnothing$. in the table~\ref{tab:m3fd_l_choice}, it is evident that the \( \ell_2 \) distance achieves the best performance. For instance, PSNR improves from 30.87 in the baseline to 31.83, while LPIPS decreases from 0.303 to 0.287, demonstrating superior reconstruction accuracy and perceptual quality. Therefore, we adopt \( \ell_2 \) distance as the evaluation function for subsequent experiments. 
Moreover, the overall performance suggests that introducing a reasonable metric function  consistently enhances the results to varying degrees. Compared to the baseline without any distance metric, these improvements indicate the strong applicability of our proposed method to different metric functions in both theoretical and practical aspects.

\begin{table}[t]
\centering
\caption{Performance and runtime comparisons on M3FD dataset.}
\vspace{-0.5em}
\fontsize{6}{7}\selectfont\setlength{\tabcolsep}{1.8mm}
\resizebox{1\columnwidth}{!}{
\begin{tabular}{l|ccccc}
\toprule
\textbf{Method} & \textbf{SSIM}~$\uparrow$ & \textbf{MSE}~$\downarrow$ & \textbf{CC}~$\uparrow$ & \textbf{LPIPS}~$\downarrow$ & \textbf{Runtime (s)} \\
\midrule
DIG-15  & 1.30 & 2771 & 0.501  & 0.3122 & 31 \\
DIG-20  & 1.38 & 2321 & 0.551  & 0.2935 & 43 \\
DIG-25  & \textbf{1.41} & \textbf{2215} & \textbf{0.5728} & \textbf{0.2870} & 52 \\
DIG-50 & 1.41 & 2219 & 0.5735 & 0.289 & 109 \\
\midrule
CCF     & 1.40 & 2271 & 0.5726 & 0.2912 & 633 \\
DDFM    & 1.40 & 2221 & 0.5684 & 0.3033 & 180 \\
\bottomrule
\end{tabular}}
\label{tab:dig_total_steps}
\vspace{-2em}
\end{table}

%% file: sec/5_Conclusion.tex
\section{Conclusion}
In this paper, we introduced a novel dynamic denoising diffusion framework for image fusion, which explicitly addresses the spatio-temporal imbalance in denoising through the lens of Diffusion Information Gains (DIG). By quantifying DIG by each modality at different noise levels, our method adaptively weights the fusion guidance to preserve critical features while ensuring high-quality, reliable fusion results. Theoretically, we proved that aligning the modality fusion weight with the corresponding guidance contribution reduces the upper bound of the generalization error, thus offering a rigorous explanation for the advantages of dynamic denoising fusion. Empirically, extensive experiments on diverse fusion scenarios, including multi-modal, multi-exposure, and multi-focus tasks, demonstrate that Dig2DIG surpasses existing diffusion-based techniques in both fusion performance and computational efficiency. As for future work, we plan to incorporate even more flexible conditioning mechanisms and generative capabilities, aiming to accommodate more complex real-world settings and further streamline the fusion pipeline. We believe that these investigations will not only broaden the scope of multi-source image fusion but also enhance the adaptability and interpretability of diffusion-based methods in challenging scenarios.